\documentstyle[ml94,epsf]{article} 

\title{{\bf Learning Word Association Norms Using 
Tree Cut Pair Models}}  
\author{Naoki Abe \hspace{0.2cm} Hang Li\\ 
Theory NEC Laboratory, 
RWCP\thanks{Real World Computing Partnership}\\ 
c/o C \& C Research Laboratories, NEC\\ 
4-1-1 Miyazaki Miyamae-ku, Kawasaki, 216 Japan. \\ 
\{abe,lihang\}@sbl.cl.nec.co.jp}

\newtheorem{proposition}{Proposition}

\newtheorem{lemma}{Lemma}

\newtheorem{definition}{Definition}

\newcommand{\tab}{\hspace{5mm}}

\begin{document}

\date{} 
\maketitle
\thispagestyle{empty}
\centerline{\large ABSTRACT} 
\bigskip 

We consider the problem of learning co-occurrence 
information between two word categories, or more in general 
between two discrete random variables taking values 
in a hierarchically classified domain. 
In particular, we consider the problem of learning the `association 
norm' defined by $A(x,y) = p(x, y)/p(x) p(y)$, where 
$p(x, y)$ is the joint distribution for $x$ and $y$ and 
$p(x)$ and $p(y)$ are marginal distributions induced by $p(x, y)$. 
We formulate this problem as a sub-task of learning 
the conditional distribution $p(x|y)$, by exploiting the 
identity $p(x|y) = A(x,y) \cdot p(x)$.  
We propose a two-step estimation method based on the MDL principle, 
which works as follows: 
It first estimates $p(x)$ as $\hat{p}$ using MDL, and then 
estimates $p(x|y)$ for a fixed $y$ by applying MDL on the hypothesis 
class of  $\{ A \cdot \hat{p} | A \in {\cal A} \}$ for some given 
class ${\cal A}$ of representations for association norm.  
The estimation of $A$ is therefore obtained as a {\em side-effect} 
of a near optimal estimation of $p(x|y)$. 
We then apply this general framework to the problem of 
acquiring case-frame patterns, an important task in corpus-based 
natural language processing.  We assume that both 
$p(x)$ and $A(x, y)$ for given $y$ are representable 
by a model based on a classification that exists within an 
existing thesaurus tree as a `cut,' and hence $p(x|y)$ is 
represented as the product of a pair of `tree cut models.'  We then 
devise an efficient algorithm that implements our general strategy. 
We tested our method by using it to actually acquire case-frame 
patterns and conducted syntactic disambiguation experiments using 
the acquired knowledge. The experimental results show that 
our method improves upon existing methods.


\section{Introduction}
\label{intro} 

A central issue in natural language processing 
is that of ambiguity resolution in syntactic parsing 
and it is generally acknowledged that  a certain amount 
of semantic knowledge is required for this.  
In particular, the case frames of verbs, namely 
the knowledge of which nouns are allowed at given 
case slots of given verbs, is crucial for this purpose.  
Such knowledge is not available in existing dictionaries 
in a satisfactory form, and hence 
the problem of automatically acquiring such knowledge 
from large corpus data has become an important topic 
in the area of natural language processing 
and machine learning. (c.f. \cite{PTL92,ALN95,LA95}) 
In this paper, we propose a new method of learning 
such knowledge, and empirically demonstrate its effectiveness. 

The knowledge of case slot patterns can be thought of 
as the co-occurrence information between verbs and 
nouns\footnote{We are interested in the co-occurrence 
information between any two word categories, but in 
much of the paper we assume that it is between nouns 
and verbs to simplify our discussion.} 
at a fixed case slot, such as at the subject position.  
In this paper, we employ the following quantity as 
a measure of co-occurrence (called `association norm'): 
\begin{equation}  
A(n, v) = \frac{p(n,v)}{p(n) p(v)}
\label{eqassoc} 
\end{equation} 
where $p(n,v)$ denotes the joint distribution over 
the nouns and the verbs (over $N \times V$), and 
$p(n)$ and $p(v)$ the marginal distributions over $N$ 
and $V$ induced by $p(n,v)$, respectively. 
Since $A(n,v)$ is obtained by dividing the joint probability of $n$ 
and $v$ by their respective marginal probabilities, 
it is intuitively clear that it measures the degree of 
co-occurrence between $n$ and $v$.  
This quantity is essentially the same as a measure proposed 
in the context of natural language processing 
by Church and Hanks \cite{CH89} called the `association ratio,'
which can be defined as $I(n,v) = \log A(n, v)$.  
Note that $I(n,v)$ is the quantity referred to as `self mutual information' 
in information theory, whose expectation with respect to $p(n,v)$ 
is the well-known `mutual information' between random 
variables $n$ and $v$.  The learning problem we are 
considering, therefore, is in fact a very general and 
important problem with many potential applications.

A question that immediately arises 
is whether the association norm as defined above 
is the right measure to use for the purpose of ambiguity resolution.  
Below we will demonstrate why this is indeed the case. 
Consider the sentence, `the sailor smacked the postman 
with a bottle.'   The ambiguity in question is between 
`smacked ... with a bottle' and `the postman with 
a bottle.'  Suppose we take the approach of comparing 
conditional probabilities, $p_{inst}(smack|bottle)$ and 
$p_{poss}(postman|bottle)$, as in some past research \cite{LA95}.    
(Here we let $p_{case}$, in general, denote the joint/conditional 
probability distribution over two word categories at the case slot 
denoted by $case$.)   
Then, since the word `smack' is such a rare word, 
it is likely that we will have $p_{inst}(smack|bottle) <
p_{poss}(postman|bottle)$, and conclude as a result that 
the `bottle' goes with the `postman.'  
Suppose on the other hand that  
we compare $A_{inst}(smack,bottle)$ and 
$A_{poss}(postman, bottle)$.  
This time we are likely to have 
$A_{inst}(smack,bottle) > A_{poss}(postman, bottle)$, 
and conclude that the `bottle' goes with `smack,' 
giving the intended reading of the sentence. 
The crucial fact here is that the two words `smack' 
and `postman' {\em have} occurred in the sentence of interest, and 
what we are interested in comparing is the respective likelihood 
that two words co-occurred at two different case slots 
(possessive/instrumental),  {\em given} that the two words 
have occurred.  It therefore 
makes sense to compare the joint probability divided 
by the respective marginal probabilities, namely 
$A(n,v) = p(n,v)/p(n) p(v)$. 

If one employed $p(n|v)$ as the measure of co-occurrence, 
its learning problem, for a fixed verb $v$, would reduce to 
that of learning a simple distribution.  In contrast, 
as $A(n,v)$ does not define a distribution, it is not immediately 
clear how we should formulate its estimation problem.  
In order to resolve this issue, we make use of 
the following identity:   
\begin{equation} 
p(n|v) = \frac{p(n,v)}{p(v)} = \frac{p(n,v)}{p(n) p(v)} \cdot p(n) 
= A(n,v) \cdot p(n).  
\label{eqid}
\end{equation} 
In other words, $p(n|v)$ can be {\em decomposed} into the product 
of the association norm and the marginal distribution over $N$. 
Now, since $p(n)$ is simply a distribution over the nouns, 
it can be estimated with an ordinary method of density estimation. 
(We let $\hat{p}(n)$ denote the result of such an estimation.) 
It is worth noting here that for this estimation, 
even when we are estimating $p(n|v)$ for a particular verb $v$, 
we can use the {\em entire} sample for $N \times V$. 
We can then estimate $p(n|v)$, using as hyopthesis class 
$H(\hat{p}) = \{ A(n,v) \cdot \hat{p}(n) | A \in {\cal A} \}$, where 
${\cal A}$ is some class of representations for the 
association norm $A(n,v)$.   Again, for a fixed verb, 
this is a simple density estimation problem, and can be 
done using any of the many well-known estimation strategies. 
In particular, we propose and employ a method based on the MDL 
(Minimum Description Length) principle \cite{Ris78,qr-idtmdlp-89}, 
thus guaranteeing a near optimal estimation of $p(n|v)$ \cite{yama}.   
As a result, we will obtain a model for $p(n|v)$, 
expressed as a product of $A(n,v)$ and $\hat{p}$, 
thus giving an estimation for the association norm $A(n,v)$ 
as a {\em side effect} of estimating $p(n|v)$. 

It has been noticed in the area of corpus-based natural language processing 
that any method that attempts to estimate either a co-occurrence 
measure or a probability value for each noun separately requires 
far too many examples to be useful in practice.  
(This is usually referred to as 
the {\em data sparseness problem}.) 
In order to circumvent this difficulty, 
we proposed in an earlier paper \cite{LA95} 
an MDL-based method that estimates $p(n|v)$   
(for a particular verb), using a noun classification 
that exists within a given thesaurus. 
That is, this method estimates the noun distribution 
in terms of a `tree cut model,' which defines a probability 
distribution by assigning a generation probability 
to each category in a `cut' within a given 
thesaurus tree.\footnote{See Section 2 for a detailed 
definition of the `tree cut models.'}
Thus, the categories in the cut are used as the `bins' of a histogram, 
so to speak.  
The use of MDL ensures that an optimal tree cut is 
selected, one that is fine enough to capture the tendency 
in the input data, but coarse enough to allow the estimation of 
probabilities of categories within it with reasonable accuracy. 
The shortcoming of the method of \cite{LA95}, however, 
is that it estimates $p(n|v)$ but {\em not} $A(n, v)$.  

In this paper, we apply the general framework of 
estimating association norm to this particular problem setting, 
and propose an efficient estimation method for $A(n,v)$ 
based on MDL. 
More formally,  we assume that the marginal distribution over 
the nouns is definable by a tree cut model, 
and that the association norm (for each verb) can also be
defined by a similar model which  
associates an $A$ value with each of the 
cateogories in a cut in the same thesaurus tree 
(called an `association tree cut model'), 
and hence $p(n|v)$ for a particular $v$ can be 
represented as the product of a pair of these tree cut models 
(called a `tree cut pair model').
(See Figure~\ref{fig:tcm}~(a),(b) and (c) 
for examples of a `tree cut,' a `tree cut model,' and 
an `association tree cut model,' all in a same thesaurus tree.)  
We have devised an efficient algorithm for 
each of the two steps in the general estimation strategy, 
namely, of finding an optimal tree cut model for the marginal 
distribution $p(n)$ (step 1), and finding an optimal association 
tree cut model for $A(n,v)$ for a particular $v$ (step 2). 
Each step will select an {\em optimal tree cut} in the thesaurus tree, 
thus providing appropriate levels of generalization 
for both $p(n)$ and $A(n,v)$. 

We tested the proposed method in an experiment, 
in which the association norms for a number of verbs and nouns 
are acquired using WordNet \cite{Miller93} as the thesaurus and 
using corpus data from the Penn Tree Bank as training data.  
We also performed ambiguity resolution experiments 
using the association norms obtained using our learning method. 
The experimental results indicate that the new method 
achieves better performance than existing 
methods for the same task, especially in terms of 
`coverage.'\footnote{Here `coverage' refers to the percentage of 
the test data for which the method could make a decision.}   
We found that the optimal tree cut found for $A(n,v)$ was 
always coarser (i.e.\ closer to the root of 
the thesaurus tree) than that for $p(n|v)$ found using 
the method of \cite{LA95}. 
This, we believe, contributes directly to the wider coverage 
achieved by our new method. 

\section{The Tree Cut Pair Model}  
\label{sec:models}

In this section, we will describe the class of representations 
we employ for distributions over nouns as well as the association 
norm between nouns and a particular verb.\footnote{In general, 
this can be between words of any two category, 
but for ease of exposition, we assume here that it is 
between nouns and verbs.}   

\begin{figure*}[tb]
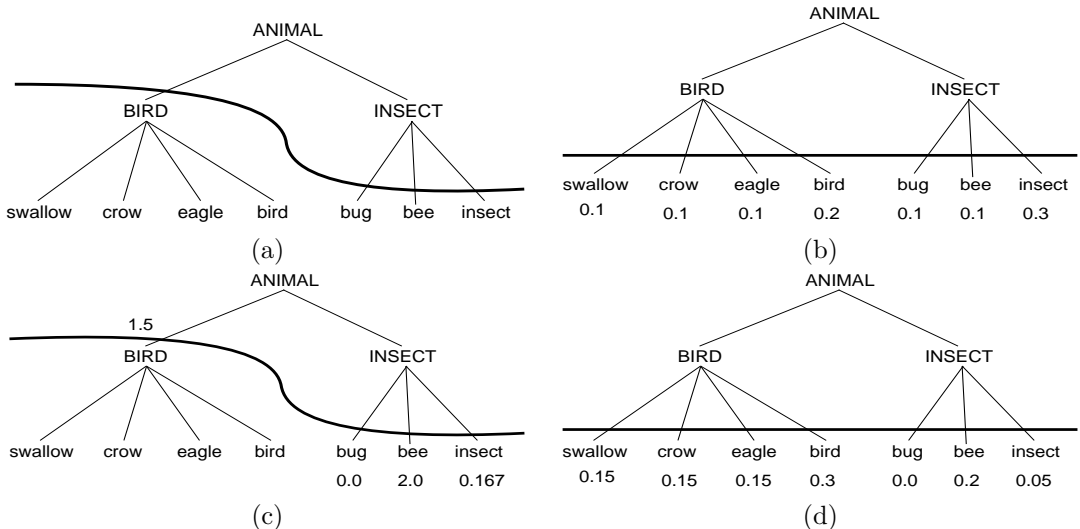

\begin{center}
$\begin{array}{cc} 
{\epsfxsize7.0cm\epsfysize2.8cm\epsfbox{treecut1.eps}}
& 
{\epsfxsize7.0cm\epsfysize3.0cm\epsfbox{tcm.eps}} \\
\mbox{(a)} & \mbox{(b)} \\ 
{\epsfxsize7.0cm\epsfysize3.0cm\epsfbox{atcm.eps}}
& 
{\epsfxsize7.0cm\epsfysize3.0cm\epsfbox{tcmp.eps}} \\
\mbox{(c)} & \mbox{(d)} 
\end{array}$  
\vspace*{-0.4cm}
\caption{(a) a tree cut (b) a TCM $p$ 
(c) an ATCM $A$ (d) distribution of $h = A \cdot p$}  
\label{fig:tcm}
\end{center}
\vspace*{-0.8cm}
\end{figure*}

A thesaurus is a tree such that each of its leaf nodes 
represents a noun, and its internal nodes represent noun 
classes.\footnote{This condition is not strictly satisfied 
by most of the publically available thesauruses, but we make 
this assumption to simplify the subsequent discussion.}
The class of nouns represented by an internal node is the set 
of nouns represented by leaf nodes dominated by that node. 
A `tree cut' in a thesaurus tree is a sequence of internal/leaf nodes, 
such that its members dominate all of the leaf nodes exhaustively 
and disjointly.  Equivalently, therefore, a tree cut is 
a set of noun categories/nouns 
which defines a partition over the set of all nouns 
represented by the leaf nodes of the thesaurus. 
Now we define the notion of a `tree cut model' (or a TCM for short)
representing a distribution over nouns.\footnote{This definition 
essentially follows that given by Li and Abe in \cite{LA95}.}  
\begin{definition}
Given a thesaurus tree $t$, 
a `tree cut model' is a pair $p = (\tau, q)$, where 
$\tau$ is a tree cut in $t$, and $q$ is a parameter vector 
specifying a probability distribution over the members of $\tau$. 
\end{definition}
A tree cut model defines a probability distribution by 
sharing the probability of each noun category uniformly 
by all the nouns belonging to that category. That is, 
the probability distribution $p$ represented by 
a tree cut model $(\tau, q)$ is given by 
\begin{equation}
\forall C \in \tau \:
\forall x \in C \: 
p(x) =  \frac{q(C)}{|C|} 
\label{deftc}
\end{equation}
A tree cut model can also be represented by a tree, 
each of whose leaf node is a pair consisting of a noun (cateogory) 
and a parameter specifying its (collective) probability. 
We give an example of a simple TCM for the category `ANIMAL' 
in Figure~\ref{fig:tcm}(b). 

We similarly define the `association tree cut model' 
(or ATCM for short). 
\begin{definition}
Given a thesaurus tree $t$ and a fixed verb $v$,  
an `association tree cut model'(ATCM) $A(\cdot, v)$ 
is a pair $(\tau, p)$, where 
$\tau$ is a tree cut in $t$, and $p$ is a function 
from $\tau$ to $\Re$. 
\end{definition}
An association tree cut model defines an association norm 
by assigning the same value $A$ of association norm to 
each noun belonging to a noun category.  That is, 
\begin{equation}
\forall C \in \tau \:
\forall x \in C \: A(x, v) =  A(C, v)
\label{defatc}
\end{equation}
We give an example ATCM in Figure~\ref{fig:tcm}(c), 
which is meant to be an ATCM for the subject slot of 
verb `fly' within the category of `ANIMAL.' 

We then define the notion of a `tree cut pair model,'  
which is a model for $p(n|v)$ for some fixed verb $v$. 
\begin{definition}
A `tree cut pair model' $h$ is a pair $h = (A, p)$,  
where $A$ is an association tree cut model (for a certain verb $v$), 
and $p$ is a tree cut model (for $N$), which satisfies the 
stochastic condition, namely, 
\begin{equation}
\sum_{n \in N} A(n, v) \cdot p(n) = 1. 
\label{eq:stoch} 
\end{equation}
\end{definition}
The above stochastic condition ensures that $h$ defines a legal 
distribution $h(n|v)$.
An example of a tree cut pair model is the pair consisting of the 
models of Figure~\ref{fig:tcm}(b) and (c), which together defines 
the distribution shown in Figure~\ref{fig:tcm}(d), verifying 
that it in fact satisfies the stochastic condition (\ref{eq:stoch}).  

\section{A New Method of Estimating Association Norms} 

As described in Introduction, our estimation procedure 
consists of two steps:  The first step is for estimating 
$p$, and the second for estimating $A$  
given an estimation $\hat{p}$ for $p$. 
The first step can be performed by an estimation method 
for tree cut models proposed by the authors in \cite{LA95}, 
and is related to `Context' of Rissanen \cite{riss83}. 
This method, called `Find-MDL,' is an efficient implementation 
of the MDL principle for the particular class of tree cut models, 
and will be exhibited for completeness, as sub-procedure of the 
entire estimation algorithm. 

Having estimated $p$ by Find-MDL using the entire sample of 
$S$ (we write $\hat{p}$ for the result of this estimation), 
we will then estimate $A$.  
As explained in Introduction, 
we will use as the hypothesis class for this estimation,  
$H(\hat{p}) = \{ A(n,v) \cdot \hat{p}(n) | A \in {\cal A}(t) \} $ 
where ${\cal A}(t)$ is the set of ATCMs for the given thesaurus tree $t$,  
and select, according to the MDL principle, 
a member of $H(\hat{p})$ that best explains the part of the 
sample that corresponds to the verb $v$, written $S_v$. That is, 
the result of the estimation, $\hat{h}$, is to be given 
by\footnote{All logarithms in this paper are to the base 2.} 
\begin{equation}
\hat{h} = \arg 
\min_{h \in H(\hat{p})} 
d.l.(h) + \sum_{n \in S_v} - \log h(n|v). 
\label{mdleq1}
\end{equation}
In the above, we used `$d.l.(h)$' to denote the model description length 
of $h$, and as is well-known, $\sum_{n \in S_v} - \log h(n|v)$ 
is the data description length for sample $S_v$ with respect to $h$. 
Since the model description length of $\hat{p}$ is fixed, 
we only need to consider the model description length of $A$,  
which consists of two parts:  
the description length for the tree cut, and that for the 
parameters. We assume that we employ the `uniform' coding scheme 
for the tree cuts, that is, all the tree cuts have exactly the 
same description length.  Thus, it suffices to consider just the 
parameter description length for the purpose of minimization.
The description length for the parameters is calculated 
as $(par(A)/2) \log |S_v|$, where $par(A)$ denotes the 
number of free parameters in the tree cut of $A$. 
Using $(1/2) \log |S_v|$ 
bits per parameter is known to be asymptotically optimal, 
since the variance of estimation is of the order $\sqrt{|S_v|}$. 
Note here that we use $\log |S_v|/2$ bits 
and {\em not} $\log |S|/2$, since the numerator $\hat{h}$ 
of $\hat{A}$ is estimated using $S_v$, 
even though the denominator $\hat{p}$  
is estimated using the entire sample $S$.  
The reason is that the estimation error for $\hat{A}$, 
provided that we assume $\hat{p}(C) \geq \epsilon$ for 
a reasonable constant $\epsilon$, is dominated by 
the estimation error for $\hat{h}$. 

Now, since we have $h(n|v) = A(n,v) \cdot \hat{p}(n)$ by definition,  
the data description length can be decomposed into the following two parts: 
\begin{equation}
\sum_{n \in S_v} - \log h(n|v) = 
\sum_{n \in S_v} - \log A(n,v) + 
\sum_{n \in S_v} - \log \hat{p}(n)
\end{equation}
Notice here that the second term does not depend on the choice 
of $A$, and hence for the minimization in (\ref{mdleq1}), it 
suffices to consider just the first term, $\sum_{n \in S_v} - 
\log A(n|v)$.  From this and the preceding discussion on 
the model description length, (\ref{mdleq1}) yields: 
\begin{equation}
\hat{h} = \arg 
\min_{h \in H(\hat{p})} 
\frac{par(A)}{2} \log |S_v| + \sum_{n \in S_v} - \log A(n,v) 
\label{mdleq2}
\end{equation}

We will now describe how we calculate the data description length 
for a tree cut pair model $h = (A, \hat{p})$. 
The data description length {\em given} a fixed tree cut is 
calculated using the maximum likelihood estimation (MLE) for $h(n|v)$, 
i.e.\ by maximizing the likelihood $L(h, S_v) = \prod_{n \in S_v} h(n|v)$. 
Since in general the tree cut of $A$ does not coincide with the 
tree cut of $\hat{p}$, this maximization problem appears somewhat involved.  
The following lemma, however, establishes that it can be 
solved efficiently.  
\begin{lemma} 
Given a tree cut model $\hat{p} = (\sigma, p)$ and 
a tree cut $\tau$, the MLE(maximum likelihood estimate) 
$\hat{h}$ = $\hat{A} \cdot \hat{p}$ is given by 
setting $\hat{h}(C'|v)$ for each $C' \in \tau$ by  
\[
\hat{h}(C'|v) = \frac{\sharp(C', S_v)}{|S_v|}  
\]
where in general we let $\sharp(C, S)$ denote the number 
of occurrences of nouns belonging to class $C$ in sample $S$. 
The estimate for $\hat{A}$ is then given by letting for each 
$C' \in \tau$,  
\[
\hat{A}(C',v) = \frac{\hat{h}(C'|v)}{\hat{p}(C')} 
\]
where $\hat{p}(C')$ is defined inductively as follows: 
\begin{enumerate}
\item 
If $C' = C$ for some $C \in \sigma$, then 
$\hat{p}(C') = \hat{p}(C)$. 
\item 
If $C'$ dominates $C_1,...,C_k$ and 
$\hat{p}(C_1),...,\hat{p}(C_k)$ are defined, then 
$\hat{p}(C') = \sum_{i=1}^{k} \hat{p}(C_i)$. 
\item 
If $C'$ is dominated by $C$ and if $\hat{p}(C)$ is defined, 
then 
$\hat{p}(C') = \frac{|C'|}{|C|} \hat{p}(C)$.
\end{enumerate}
\label{lem:mle}
\end{lemma} 

\noindent{{\bf Proof of Lemma~\ref{lem:mle}}}

Given the tree cuts, $\tau$ and $\sigma$, define 
$\tau \wedge \sigma$ to be the tree cut whose 
noun partition equals the coarsest partition 
that is finer than or equal to both the noun partitions 
of $\tau$ and $\sigma$. 
Then, the likelihood function $L(h, S_v)$ 
which we are trying to maximize (for $h = (A, \hat{p})$)
can be written as follows,  
\begin{equation}
L(h, S_v) = \prod_{C \in \tau \wedge \sigma} 
(A(C, v) \cdot \hat{p}(C))^{\sharp(C, S_v)}     
\label{eq:max1}
\end{equation}
where $A(C|v)$ for $C \not\in \tau$ and 
$\hat{p}(C)$ for $C \not\in \sigma$ are defined so that 
they be consistent\footnote{That is, 
$\hat{p}$ is defined as specified in the lemma, and 
$A$ is defined by inheriting the same value as the $A$ value 
of the ascendant in $\tau$.} 
with the definitions of $A(n|v)$ and $\hat{p}(n)$. 
As before, since $\hat{p}$ is fixed, 
the above maximization problem is equivalent to maximizing 
just the product of $A$ values, namely,  
\begin{equation}
\arg \max_A L(h, S_v) = 
\arg \max_A \prod_{C \in \tau \wedge \sigma} A(C, v)^{\sharp(C, S_v)} 
\label{eq:max2}
\end{equation}
Since $\tau \wedge \sigma$ is always finer than $\tau$, 
for each $C \in \tau \wedge \sigma$, 
there exists some $C' \in \tau$ such that  
$A(C, v) = A(C', v)$. Thus, 
\begin{equation}
\arg \max_A L(h, S_v) = 
\arg \max_A \prod_{C' \in \tau} A(C', v)^{\sharp(C', S_v)} 
\label{eq:max3}
\end{equation}
Note that the maximization is subject to the condition:  
\begin{equation}
\sum_{n \in S_v} A(n,v) \cdot \hat{p}(n) 
= \sum_{C' \in \tau} A(C',v) \cdot \hat{p}(C') 
= 1. 
\label{eq:cond}
\end{equation}
Since multiplying by a constant leaves 
the argument of maximization unchanged, (\ref{eq:max3}) yields 
\begin{equation}
\arg \max_A L(h, S_v) = 
\arg \max_{A} \prod_{C' \in \tau} 
(A(C', v) \cdot \hat{p}(C'))^{\sharp(C', S_v)}
\label{eq:max4}
\end{equation}
where the maximization is under the same condition (\ref{eq:cond}). 
Emphatically, the quantity being maximized in (\ref{eq:max4}) 
is {\em different} from the likelihood in (\ref{eq:max1}), 
but both attain maximum for the same values of $A$.  
Thus the maximization problem is reduced to one of the form: 
`maximize $\prod (a_i \cdot p_i)^{k_i}$ subject to 
$\sum a_i \cdot p_i = 1$.  
As is well-known, this is given by setting, 
$a_i \cdot p_i = \frac{k_i}{\sum_i k_i}$ for each $i$. 
Thus, (\ref{eq:max2}) is obtained by setting, for 
each $C' \in \tau$, 
\begin{equation}
h(C') = A(C', v) \cdot \hat{p}(C') 
= \frac{\sharp(C', S_v)}{|S_v|}
\end{equation}
Hence, $\hat{A}$ is given, for each $C' \in \tau$, by 
\begin{equation}
\hat{A}(C', v) = \frac{\hat{h}(C'|v)}{\hat{p}(C')}  
\end{equation}
This completes the proof.  $\Box$  

We now go on to the issue of how 
we can find a model satisfying (\ref{mdleq2}) efficiently:  
This is possible with a recursive algorithm which 
resembles Find-MDL of \cite{LA95}. 
This algorithm works by recursively applying itself on 
subtrees to obtain optimal tree cuts for each of them, and 
decides whether to return a tree cut obtained by appending 
all of them, or a cut consisting solely of the top node of the 
current subtree, by comparison of the respective description length. 
In calculating the data description length at each recursive call, 
the formulas of Lemma~\ref{lem:mle} are used to obtain the MLE. 
The details of this procedure are shown below 
as algorithm `Assoc-MDL.'   
Note, in the algorithm description, that $S$ denotes the input 
sample, which is a sequence of elements of $N \times V$.  
For any fixed verb $v \in V$, 
$S_v$ denotes the part of $S$ that corresponds to verb $v$, 
i.e. $S_v= \{ n \in S | (n,v) \in S \}_M$. 
(We use $\{ \}_M$ when denoting a `multi-set.')
We use $\pi_1(S)$ to denote the multi-set of nouns 
appearing in sample $S$, i.e. 
$\pi_1(S) = \{ n \in S | \exists v \in V \: (n,v) \in S \}_M$. 
In general, $t$ stands for a node in a tree, or equivalently 
the class of nouns it represents. 
It is initially set to the root node of the input thesaurus tree.  
In general, `$[...]$' denotes a list. 
\begin{tabbing}
{\bf algorithm} Assoc-MDL($t, S$) \\ 
1.  $\hat{p}$ := Find-MDL($t, \pi_1(S)$) \\ 
2.  $\hat{A}$ := Find-Assoc-MDL($S_v, t, \hat{p}$)\\ 
3.  return(($\hat{A},\hat{p}$)) \\ 
\\
{\bf sub-procedure} Find-MDL($t, S$) \\ 
1.  {\bf if} $t$ is a leaf node \\ 
2.  {\bf then} return($([t], \hat{p}(t, S))$) \\ 
3.  {\bf else} \\
4.  \tab For each child $t_i$ of $t$,   
$c_i$ $:=$Find-MDL($t_i, S$)\\ 
5.  \tab $\gamma$$:=$ append($c_i$) \\ 
6.  \tab {\bf if} 
$\sharp(t, \pi_1(S)) (- \log \frac{\hat{p}(t)}{|t|})   
+ \frac{1}{2} \log N$ $<$  \\ 
 \tab \tab  
$\sum_{t_i \in children(t)} \sharp(t_i, \pi_1(S)) 
(- \log \frac{\hat{p}(t_i)}{|t_i|}) + \frac{|\gamma|}{2} \log N$ \\ 
7.  \tab {\bf then} return($([t], \hat{p}(t, S))$) \\ 
8.  \tab {\bf else} return$(\gamma)$ \\ 
\end{tabbing}
\begin{tabbing}
{\bf sub-procedure} Find-Assoc-MDL($S_v, t, \hat{p}$) \\ 
1.  {\bf if} $t$ is a leaf node \\ 
2.  {\bf then} return($([t], \hat{A}(t,v))$) \\ 
3.  {\bf else} Let $\tau :=$ {\em children}$(t)$ \\ 
4.  \tab $\hat{h}(t|v) := \frac{\sharp(t,S_v)}{|S_v|}$\\
5.  \tab $\hat{A}(t,v) := 
\frac{\hat{h}(t|v)}{\hat{p}(t)}$\\
 /* We use definitions in Lemma~\ref{lem:mle} 
to calculate $\hat{p}(t)$ */\\
6.  \tab For each child $t_i \in \tau$ of $t$ \\ 
7.  \tab $\gamma_i$ $:=$Find-Assoc-MDL($S_v, t_i, \hat{p}$)\\ 
8.  \tab $\gamma$$:=$ append($\gamma_i$) \\ 
9.  \tab {\bf if} 
$ \sharp(t, S_v) (- \log \hat{A}(t, v)) + \frac{1}{2} \log |S_v|$ $<$ \\ 
\tab \tab 
$\sum_{t_i \in \tau} \sharp(t, S_v) (- \log \hat{A}(t_i,v)) 
+ \frac{|\tau|}{2} \log |S_v|$ \\ 
 /* The values of $\hat{A}(t_i,v)$ used above are to be */ \\ 
 /* those in $\gamma_i$ */ \\ 
11. {\bf then} return($([t], \hat{A}(t, v))$) \\ 
12. {\bf else} return($\gamma$)
\end{tabbing}

Given Lemma~\ref{lem:mle}, it is not difficult to see that 
Find-Assoc-MDL indeed does find a tree cut pair model 
which minimizes the total description length. 
Also, its running time is clearly linear in the size
(number of leaf nodes) in the thesaurus tree, 
and linear in the input sample size. 
The following proposition summarizes these observations. 

\begin{proposition}
Algorithm {\em Find-Assoc-MDL} outputs 
$\hat{h} \in H(\hat{p}) = \{ A \cdot \hat{p} | A \in {\cal A}(t) \}$ 
(where ${\cal A}(t)$ denotes the class of association tree cut 
models for thesaurus tree $t$)  
such that 
\[
\hat{h} = \arg \min_{h \in H(\hat{p})} d.l.(h) 
+ \sum_{n \in S_v} - \log h(n|v) 
\]
and its worst case running time is $O(|S| \cdot |t|)$,  
where $|S|$ is the size of the input sample, and 
$|t|$ is the size (number of leaves) of the thesaurus tree. 
\end{proposition}
We note that an analogous (and easier) proposition on Find-MDL 
is stated in \cite{LA95}.  

\section{Comparison with Existing Methods}

A simpler alternative formulation of the problem of acquiring 
case frame patterns is to think of it 
as the problem of learning the distribution over 
nouns at a given case slot of a given verb, as in \cite{LA95}.  
In that paper, the algorithm Find-MDL 
was used to estimate $p(n|v)$ for a fixed verb $v$, 
which is merely a distribution over nouns. 
The method was guaranteed to be near-optimal as a method of 
estimating the noun distribution, but it suffered from the 
disadvantage that it tended to be influenced by the {\em absolute 
frequencies} of the nouns.  This is a direct consequence 
of employing a simpler formulation of the problem, 
namely as that of learning a distribution over nouns at a given case 
slot of a given verb, and {\em not} an association norm 
between the nouns and verbs.

To illustrate this difficulty, suppose that 
we are given 4 occurrences of the word `swallow,' 
7 occurrences of `crow,' and 1 occurrence of `robin,' 
say at the subject position of `fly.'  
The method of \cite{LA95} would probably conclude that 
`swallow' and `crow' are likely to appear 
at subject position of `fly,' but not `robin.'  
But, the reason why the word `robin' is not observed many times 
may be attributable to the fact that this word simply  
has a low absolute frequency, irrespective of the context. 
For example, `swallow,' `crow,' and `robin' might each have 
absolute frequencies of 42, 66, and 9, in the same data 
with unrestricted contexts.  
In this case, their frequencies of 4, 7 and 1 as subject of `fly' 
would probably suggest that they are all roughly equally likely 
to appear as subject of `fly,' given that they do appear at all. 

An earlier method proposed by Resnik \cite{Res92} takes into 
account the above intuition in the form of a heuristic. 
His method judges whether a given noun tends to 
co-occur with a verb or not, based on its super-concept 
having the highest value of association norm with that verb.  
The association norm he used, called the `selectional 
association' is defined, for a noun class $C$ and a verb $v$, 
as  
\[ 
\sum_{n \in C} p(n) \log \frac{p(n,v)}{p(n) p(v)}. 
\]
Despite its intuitive appeal, 
the most serious disadvantage of Resnik's method, in our view,   
is the fact that no theoretical justification is provided for 
employing it as an estimation method, in contrast to 
the method of Li and Abe \cite{LA95}, which enjoyed theoretical 
justification, if at the cost of an over-simplied formulation. 
It thus naturally leads to the question of whether there exists 
a method which estimates a reasonable notion of association norm,
and at the same time is theoretically justified as an estimation
method.  This, we believe, is exactly what the method 
proposed in the current paper provides. 

\section{Experimental Results}

\vspace*{-0.2cm} 
\subsection{Learning Word Assocation Norm} 
\vspace*{-0.1cm} 

\begin{figure*}[tb]
\begin{center}
{\epsfxsize14.0cm\epsfysize3.8cm\epsfbox{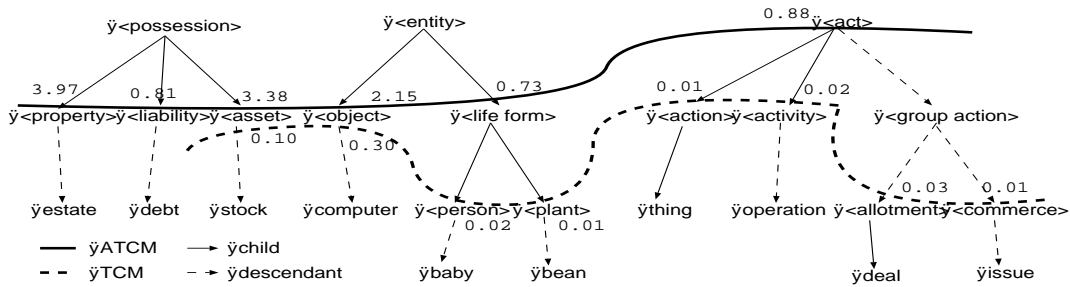}}
\end{center}
\caption{Parts of the ATCM and the TCM} 
\label{fig:wordnet} 
\vspace*{-0.5cm}
\end{figure*}

The training data we used were obtained from the texts of the {\em
  tagged} Wall Street Journal corpus (ACL/DCI CD-ROM1), which contains
126,084 sentences.  In particular, we extracted triples of the form
$(verb, case\_slot, noun)$ or $(noun, case\_slot, noun)$ using a
standard pattern matching technique.  (These two types of triples can
be regarded more generally as instances of $(head, case\_slot,
slot\_value)$.)  The thesaurus we used is basically `WordNet'
(version1.4) \cite{Miller93}, but as WordNet has some anomalies which
make it deviate from the definition of a `thesaurus tree' we had in
Section~\ref{sec:models}, we needed to modify it
somewhat.\footnote{These anomalies are: (i) The structure of WordNet
  is in fact not a tree but a DAG; (ii) The (leaf and internal) nodes
  stand for a word sense and not a word, and thus the same word can be
  contained in more than one word senses and vice-versa.  We refer the
  interested reader to \cite{LA95} for the modifications we made.}
Figure~\ref{fig:wordnet} shows selected parts of the ATCM obtained by
Assoc-MDL for the direct object slot of the verb `buy,' as well as the
TCM obtained by the method of \cite{LA95}, i.e.\ by applying Find-MDL
on the data for that case slot.  Note that the nodes in the TCM having
probabilities less than 0.01 have been discarded.

We list a number of general tendencies 
that can be observed in these results. 
First, many of the nodes that 
are assigned high $A$ values by the ATCM are not present in the 
TCM, as they have negligible absolute frequencies. 
Some examples of these nodes are $\langle property, belonging... \rangle$, 
$\langle right \rangle$, $\langle owernership \rangle$, and 
$\langle part,... \rangle$. 
Our intuition agrees with the judgement that they do 
represent suitable direct objects of `buy,' and the fact 
that they were picked up by Assoc-MDL despite their 
low absolute frequencies seems to confirm the advantage of 
our method. 
Another notable fact is that the cut in the ATCM is 
always `above' that of the TCM. 
For example, as we can see in Figure~\ref{fig:wordnet}, 
the four nodes $\langle action \rangle$, $\langle activity \rangle$, $\langle allotment \rangle$, 
and $\langle commerce \rangle$ in the TCM are all generalized as 
one node $\langle act \rangle$ in the ATCM, reflecting the judgement 
that despite their varying absolute frequencies, their association 
norms with `buy' do not significantly deviate from one another. 
In contrast, note that the nodes 
$\langle property \rangle$, $\langle asset \rangle$, and $\langle liability \rangle$ are kept 
separate in the ATCM, as the first two have high $A$ values, 
whereas $\langle liability \rangle$ has a low $A$ value, which is 
consistent with our intuition that one does not want to buy debt. 

\vspace*{-0.1cm} 
\subsection{PP-attachment Disambiguation Experiment}
\vspace*{-0.1cm} 

We used the knowledge of association norms acquired in the 
experiment described above to resolve pp-attachment ambiguities.  

For this experiment, we 
used the bracketed corpus of the Penn Tree Bank (Wall Street
Journal Corpus) \cite{Marcus93} as our data.  First we randomly 
selected one directory of the WSJ files containing roughly
$1/26$ of the entire data as our test data and what remains as the
training data. We repeated this process ten times 
to conduct {\em cross validation}.
At each of the ten iterations, we extracted from the test data 
$(verb,noun_1,prep,noun_2)$ quadruples, as well as the 
`answer' for the pp-attachment site for each quadruple by 
inspecting the parse trees given in the Penn Tree Bank.  
Then we extracted from the training data 
$(verb,prep,noun_2)$ and $(noun_1,prep,noun_2)$ triples. 
Having done so, we preprocessed both the training and test data 
by removing obviously noisy examples, and subsequently applying 
12 heuristic rules, including: 
(1) changing the inflected form of a word to its stem form,
(2) replacing numerals with the word `number,' (3) replacing integers
between $1900$ and $2999$ with the word `year,' etc..\ 
On the average, for each iteration we obtained $820.4$ quadruples 
as test data, and $19739.2$ triples as training data.

For the sake of comparison, we also tested the method proposed 
in \cite{LA95}, as well as a method based on Resnik's \cite{Res92}. 
For the former, we used Find-MDL to learn the distribution of 
$case\_value$s (nouns) at a specific $case\_slot$ 
of a specific $head$ (a noun or a verb), 
and used the acquired conditional probability distribution 
$p_{head}(case\_value | case\_slot)$ to disambiguate the test patterns.  
For the latter, we generalized each $case\_value$ at a 
specific $case\_slot$ of a specific $head$ to the appropriate level 
in WordNet 
using the `selectional association' (SA) measure, and used the 
SA values of those generalized classes for 
disambiguation.\footnote{Resnik actually generalizes both the 
$head$s and the $case\_value$s, but here we only generalize 
$case_value$s to allow a fair comparison.} 

More concretely, for a given test pattern 
($verb$, $noun_1$, $prep$, $noun_2$), 
our method compares $\hat{A}_{prep}$$(noun_2, verb)$ 
and $\hat{A}_{prep}$$(noun_2, noun_1)$, 
and attach $(prep, noun_2)$ to $verb$ or $noun_1$ 
depending on which is larger. 
If they are equal, then it is judged that no decision can be made. 
Disambiguation using SA is done in a similar manner, by comparing 
the two corresponding SA values, while that by Find-MDL 
is done by comparing the conditional probabilities, 
$\hat{P}_{prep}(noun_2|$ $verb$) 
and $\hat{P}_{prep}(noun_2|$ $noun_1$). 

Table~\ref{tab:results} shows the results of this pp-attachment
disambiguation experiment in terms of `coverage' and `accuracy.' 
Here `coverage' refers to the percetage of the test patterns 
for which the disambiguation method made a decision, 
and `accuracy' refers to the percetage of those decisions 
that were correct.  In the table, `Default' refers to the method 
of always attaching $(prep,noun_2)$ to $noun_1$, and 
`Assoc' `SA,' and `MDL' stand for using Assoc-MDL,  
selectional association, and Find-MDL, respectively.  
The tendency of these results is clear: 
In terms of prediction accuracy, 
Assoc remains essentially unchanged from both SA and MDL 
at about 95 per cent.  In terms of coverage, however, 
Assoc, at 80.0 per cent, significantly out-performs both SA and MDL, 
which are at 63.7 per cent and 73.3 per cent, respectively.

\begin{table} 
\begin{center}
\begin{tabular}{|l|c|c|} \hline
 & Coverage(\%) & Accuracy(\%) \\ \hline
Default & $100$ & $70.2$ \\
MDL & $73.3$ & $94.6$ \\ 
SA & $63.7$ & $94.3$ \\ 
Assoc & $80.0$ & $95.2$ \\ \hline 
\end{tabular}
\end{center}
\caption{Results of PP-attachment disambiguation} 
\label{tab:results} 
\end{table}

\begin{figure}[tb]
\begin{center}
{\epsfxsize8.2cm\epsfysize5.0cm\epsfbox{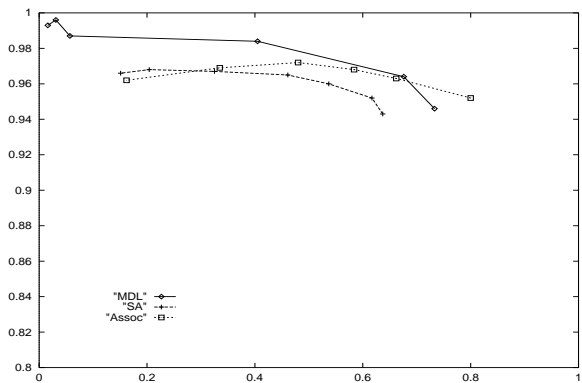}}
\caption{The coverage-accuracy curves for MDL, SA and Assoc.}
\label{fig:curves}
\end{center}
\vspace*{-0.5cm} 
\end{figure}

Figure~\ref{fig:curves} plots the `coverage-accuracy' curves 
for all three methods.  The x-axis is the coverage 
(in ratio not in percentage) and the y-axis is the accuracy. 
These curves are obtained by employing a 
`confidence test'\footnote{We perform 
the following heuristic confidence test to judge whether a 
decision can be made.  We divide the difference between the 
two estimates by the approximate standard deviation 
of that difference, heuristically calculated by 
$\sqrt{\frac{\hat{\sigma}_1^2}{N_1} +
  \frac{\hat{\sigma}_2^2}{N_2}}$, where $\hat{\sigma}_i^2$ is the
variance of the association values for the classes in 
the tree cut output for $head$ and $prep$ in question, 
and $N_i$ is the size of the corresponding sub-sample.
(The test is simpler for MDL.) } 
for judging whether to make a decision or not, and then 
changing the threshold confidence level as parameter. 
It can be seen that overall Assoc enjoys a higher coverage 
than the other two methods, since its accuracy does not 
drop nearly as sharply as the other two methods   
as the required confidence level approaches zero. 
Note that ultimately what matters the most is the performance 
at the `break-even' point, namely the point at which the 
accuracy equals the coverage, since it achieves the optimal 
accuracy overall. 
It is quite clear from these curves that Assoc will win out there. 
The fact that Assoc appears to do better than MDL confirms our intuition 
that the association norm is better suited for the purpose of 
disambiguation than the conditional probability.  
The fact that Assoc out-performs SA, on the other hand, 
confirms that our {\em estimation method} for 
the association norm based on MDL is not only theoretically 
sound but excels in practice, as SA is a heuristic method 
based on essentially the same notion of association norm.  


\section{Concluding Remarks} 

We have proposed a new method of learning the `association norm'
$A(x,y) = p(x,y)/p(x) p(y)$ between two discrete random variables.  We
applied our method on the important problem of learning word
association norms from large corpus data, using the class of `tree cut
pair models' as the knowledge representation language.  A syntactic
disambiguation experiment conducted using the acquried knowledge shows
that our method improves upon other methods known in the literature
for the same task.  In the future, we hope to demonstrate that the
proposed method can be used in practice, by testing it on even larger
corpus data.

\vspace*{-0.2cm}
\section*{Acknowledgement}
\vspace*{-0.1cm}

We thank Ms.\ Y.\ Yamaguchi of NIS for her programming efforts, 
and Mr.\ K.\ Nakamura and Mr.\ T.\ Fujita 
of NEC Corporation for their encouragement. 

\newcommand{\etalchar}[1]{$^{#1}$}

\end{document}